\pdfoutput=1
\documentclass[11pt]{article}

\usepackage[]{EACL2023}

\usepackage{times}
\usepackage{latexsym}

\usepackage[T1]{fontenc}
\usepackage[utf8]{inputenc}

\usepackage{microtype}

\usepackage{inconsolata}
\usepackage{comment} %by SP
\usepackage{todonotes} %by SP
\usepackage{caption,subcaption} % SP
\usepackage[normalem]{ulem} %SP
\useunder{\uline}{\ul}{} %SP

\title{A Psycholinguistic Analysis of BERT's Representations of Compounds}

\author{Lars Buijtelaar \\
  University of Amsterdam\\
  \texttt{lars.buijtelaar@student.uva.nl} \\\And
  Sandro Pezzelle \\
  ILLC, University of Amsterdam\\
  \texttt{s.pezzelle@uva.nl} \\}

\begin{document}

\maketitle
\begin{abstract}
This work studies the semantic representations learned by BERT for compounds, that is, expressions such as \emph{sunlight} or \emph{bodyguard}. We build on recent studies that explore 
semantic information in Transformers at the word level 
and test whether BERT aligns with human semantic intuitions when dealing with expressions (e.g., \emph{sunlight}) whose overall meaning depends---to a various extent---on the semantics of the constituent words (\emph{sun}, \emph{light}). We leverage a dataset that includes human judgments on two psycholinguistic measures of compound semantic analysis: lexeme meaning dominance (LMD; quantifying the weight of each constituent toward the compound meaning) and semantic transparency (ST; evaluating the extent to which the compound meaning is recoverable from the constituents' semantics). We show that BERT-based measures moderately align with human intuitions, especially when using contextualized representations, and that LMD is overall more predictable than ST. Contrary to the results reported 
for `standard' words,
higher, more contextualized layers are the best at representing compound meaning.
These findings shed new light on 
the abilities of BERT in dealing with 
fine-grained semantic phenomena. 
Moreover, they can provide insights into 
how speakers represent 
compounds.

\end{abstract}

\section{Introduction}

Compounds such as \emph{sunlight} or \emph{bodyguard} are an interesting benchmark to probe 
the semantic representations learned by any NLP models.
On the one hand, compounds that are part of a language lexicon~\cite[i.e., lexicalized compounds;][]{gagne2006} have their own (sets of) established meaning(s). As such, they are lexical items just like any other word.
On the other hand, the semantic status of compounds is special since their meaning is the result of the combination of the meaning of two words (hence, the constituents). According to psycholinguistic evidence, this semantic relation does not disappear 
with
lexicalization. Indeed, speakers 
actively combine constituent meanings when processing both 
novel and lexicalized compounds~\cite{gagnesp2009,ji2011,marelli2012,marelli2014}.

In this work, we argue that NLP systems capable of faithfully representing word meanings should account for these aspects. For example, to acknowledge that the meaning of \emph{handgun} relies more on the semantics of \emph{gun} than of \emph{hand} (indeed, a \emph{handgun} is a type of \emph{gun}). Or, that the meaning of \emph{sunlight} is more directly recoverable from the semantics of its constituents (it is more transparent) than is the meaning of \emph{muskrat} (which is very opaque).

Transformer-based encoders such as BERT~\cite{devlin2019bert} 
are shown
to produce word representations that align well with human semantic intuitions, particularly at their 
lower layers~\cite{bommasani2020interpreting,vulic-etal-2020-probing}. This suggests that these models are effective in 
encoding the meaning of a word, 
without any additional task-specific fine-tuning. However,
these conclusions are based on evaluations that explore 
semantic
relations 
between words, such as pairwise similarity patterns---not between words and their parts.

In parallel, BERT's contextualized embeddings have been leveraged for tasks that involve lexical composition. For example, to learn
compound representations that are effective in predicting 
the \emph{literality} of a compound or its semantic interpretation~\cite{10.1162/tacl_a_00277}.
In this case, mixed results were reported. While BERT-based models are effective to judge, e.g.,
that \emph{market} (but not \emph{flea}) has a literal meaning in \emph{flea market}, 
they are
far behind humans in predicting, e.g., that \emph{body part} stands for \emph{part that makes up a body}. Crucially, these results were obtained by training a binary classifier on the top of BERT's embeddings. Since the encoder parameters were updated during learning,
no conclusions can be drawn on 
the effectiveness of BERT's embeddings in dealing with these and similar fine-grained semantic aspects.

In this work, we leverage a dataset that includes human judgments on two psycholinguistic measures of compound semantic analysis: lexeme meaning dominance (LMD) and semantic transparency (ST). The former quantifies the semantic weight of each constituent toward the compound meaning. For example, that \emph{gun} has more semantic weight in \emph{handgun} than \emph{hand} does. The second evaluates the extent to which the compound meaning is recoverable from the semantics of the constituents. For example, that \emph{handgun} is very transparent, while \emph{muskrat} is much less so.

We 
test 
whether, and to what extent, the measures of LMD and ST that we obtain from BERT's representations of compounds and compound constituents 
align with human judgments.
We carry out comprehensive experiments on model versions, contexts, pooling methods, layers.\footnote{Data and code are made available at \url{https://github.com/lars927/compounds-analysis-bert}} We show that:

\begin{itemize}
    \item  BERT is moderately aligned with human intuitions on both measures, which confirms the effectiveness of the model in accounting for the fine-grained semantic aspects captured by LMD and ST. At the same time, LMD is substantially more predictable than ST;
    
    \item  only representations extracted from words in a context (in a sentence), but not without a context (in isolation), are aligned with human intuitions, which reflects BERT's struggle to handle out-of-context words. Moreover, 
    the highest correlations are achieved in higher, deeply contextualized layers. 
    This could be due to the nature of the semantic evaluation subtending LMD and ST, which likely requires relying on a specific semantic interpretation of the compound rather than on abstract lexico-semantic information;

    \item  both BERT\textsubscript{base} and BERT\textsubscript{large} outperform the GloVe~\cite{pennington2014glove} baseline, with BERT\textsubscript{large} achieving the best results overall. This confirms the effectiveness of BERT models to represent word-level semantics, in line with previous work~\cite{bommasani2020interpreting,vulic-etal-2020-probing};

    \item BERT accounts for the left and right constituents equally when representing the semantics of a compound.
    Moreover, these representations appear to 
    encode the complex semantic and syntactic relation tying the constituents.
    
\end{itemize}

\section{Related Work}

\subsection{Compound Semantics in Psycholinguistics}

Compounds are one of the favorite subjects of psycholinguistic research. One of the reasons is that they are extremely productive: a new combination of two (or more) words can be generated at any time and get lexicalized through language use~\cite{gagne2006}. Indeed, compounds have been considered to serve as a ``backdoor into the lexicon''~\cite{downing1977}.
While understanding novel compounds clearly involves accessing both the meaning of the constituents and the semantic relation tying them together, recent
psycholinguistic evidence has shown that an active combination of the meaning of the constituent words is routinely in place also for lexicalized compounds~\cite{gagnesp2009,ji2011,marelli2012,marelli2014}. 
Indeed, most psycholinguistic research in this field focuses on the constituents and their relation with compounds. For example, to study and quantify the role of frequency, semantic transparency, or headedness~\cite{gagnesp2009,marelli2009head,marelli2012,juhasz2015database}.\footnote{Headedness refers to the property of having a head---in English, typically the right constituent. The left is the modifier.}

Recently, a few studies leveraging methods from NLP have been carried out to either reproduce or quantify some of these aspects. By typically using \emph{static} embeddings~\cite{mikolov2013,pennington2014glove} and compositional models of distributional semantics~\cite{mitchell2010,guevara2010}, 
these approaches have proven successful
in building compound representations
that approximate, e.g., the different semantic and syntactic role of a compound's modifier and head, semantic transparency, plausibility of a novel combination or syntax-based categorizations~\cite{gunther2016understanding,marelli2017compounding,gunther2019enter,pezzelle2020semantic}.

Though powerful, these methods have one crucial limitation, namely, they require training a set of parameters via supervised learning to obtain representations for compounds that are novel or simply not present in the corpus. Transformer-based encoders such as BERT~\cite{devlin2019bert} lift this limitation. Without any additional training or fine-tuning, in fact, they can represent any novel or unseen word---provided that it can be divided into known subwords. Since compounds involve a meaningful combination of two constituents, they represent an interesting benchmark to test the representations by these models.

\subsection{Word Representation in Transformers}

A recent line of work started to investigate 
the type of semantic information encoded in the embeddings 
by pre-trained Transformer encoders.
While this was a classical benchmark to evaluate static, word-level embeddings learned by previous generation models, e.g., word2vec~\cite{mikolov2013} or GloVe~\cite{pennington2014glove}, the
problem appears less trivial for current state-of-the-art NLP models~\cite{westera2019don,mickus2020you,lenci2022comparative}. Indeed, the embeddings by Transformer-based encoders are \emph{contextualized}, i.e., affected by both the surrounding context and the position within a sentence. Moreover, they often represent subwords rather than whole words.

By means of a simple method to pool the various contextualized embeddings learned for a word into a single, static embedding,~\citet{bommasani2020interpreting} showed that these representations 
align with human judgments of semantic similarity better than how previous-generation ones do.
In particular, lower layers perform the best, which reveals that these layers encode abstract, lexico-semantic information.
Similar findings were reported by~\citet{vulic-etal-2020-probing}, who extended the investigation to other five languages than English. Taken together, these results are complementary to the findings that representations in higher layers tend to become more context-specific~\cite{ethayarajh-2019-contextual} and 
to better encode word senses~\cite{reif2019visualizing}.

Recent work~\cite{10.1162/tacl_a_00277} leveraged BERT embeddings to obtain representations for compounds by means of (trained) lexical composition models---similarly to how it was done for compositional distributional semantic models. However, to the best of our knowledge, no work to date has explored how Transformer-based encoders represent compounds. 
The most relevant study in this direction is the one by~\citet{pinter-etal-2020-will}, which focused on BERT's representations for blends (i.e., words such as \emph{shoptics}, resulting from the merging of \emph{shop} and \emph{optics}) and included a comparison with novel compounds. They 
reported an overall high similarity between the compound and the constituents, slightly increasing through the layers.

By focusing on lexicalized compounds
from a psycholinguistic angle,
we are the first to study how BERT represents these complex expressions.

\begin{table}[]
\centering
\begin{tabular}{lcc}
\hline
\textbf{compound} & \textbf{LMD {[}0,10{]}} & \textbf{ST {[}1,7{]}} \\ \hline
handgun                           & 8.13    $\rightarrow$      & 6.29  $\uparrow$          \\ 
bodyguard                         & 7.27     $\rightarrow$  & 5.64 $\uparrow$              \\
%\hline
policeman                         & 3.07  $\leftarrow$        & 6.13 $\uparrow$           \\
wartime                            & 3.47 $\leftarrow$         & 6.31 $\uparrow$           \\ %\hline
muskrat                            & 7.53 $\rightarrow$     & 2.80 $\downarrow$              \\
primrose                           & 7.93  $\rightarrow$       & 2.00  $\downarrow$           \\ %\hline
milestone                           & 3.36      $\leftarrow$      & 2.21 $\downarrow$         \\
cheapskate                         & 2.00   $\leftarrow$     & 2.00  $\downarrow$             \\ \hline
\end{tabular}
\caption{A few examples from the dataset with either high $\uparrow$ or low $\downarrow$ ST and either low $\leftarrow$ or high $\rightarrow$ LMD. E.g., the meaning of \emph{handgun} is deemed highly transparent 
% (6.29) 
and based more on the right than the left constituent.}\label{tab:stats}
% (8.13).}
\end{table}

\section{Data}

We use a psycholinguistic dataset of human judgments on compound LMD and ST~\cite{juhasz2015database}. The dataset includes 629 lexicalized English compounds annotated by 189 participants for 
various variables.\footnote{Such as LMD, ST,
age of acquisition, and imageability.} LMD is a score that captures which of the two constituents of a compound is semantically dominant for the compound meaning. It ranges in [0,10], where 0 means totally dependent on the left constituent and 10 means totally dependent on the right constituent. In Table~\ref{tab:stats}, we report a few examples from the dataset. As can be seen, compounds such as \emph{handgun} or \emph{muskrat} have a high LMD, i.e., the right constituent is semantically dominant. In contrast, compounds such as \emph{policeman} or \emph{milestone} have a low LMD, i.e., the left 
constituent 
is semantically dominant. 

ST is defined as a score that quantifies the degree to which the meaning of a compound can be inferred or recovered from the meaning of the constituents: the higher the ST, the more transparent the compound. 
The compounds \emph{handgun} and \emph{wartime} in Table~\ref{tab:stats}, for example, are fully transparent: both the constituents contribute to their meaning. 
In contrast, 
compounds such as \emph{primrose} or \emph{cheapskate}
are fully opaque: neither of the two constituents contributes to its meaning.

Since only the compounds, but not the constituents, are provided in the dataset, we manually annotate each compound (e.g., \emph{handgun}) with its left (\emph{hand}) and right (\emph{gun}) constituents, so to obtain a dataset of $\langle$compound, left, right$\rangle$ triplets. While doing so, we decided to discard the pseudo-compound \emph{mushroom}. We were left with 628 triplets, that we use in our experiments.

\section{Method}

We test whether,
and to what extent, BERT's representations of compounds and compound constituents 
approximate human judgments on LMD and ST. To do so, we obtain word-level representations using two versions of BERT.
We experiment with representations obtained by feeding the word either in isolation or in the context of a sentence. 
Moreover, building on previous work, we explore various pooling methods 
over BERT outputs.

\subsection{Models}

BERT~\cite{devlin2019bert} is a Transformer-based model pre-trained on a large number of English texts. It is pre-trained using two learning objectives, i.e., Masked Language Modeling (MLM) and Next Sentence Prediction (NSP). MLM is about predicting some words that have been masked in the input. NSP is about predicting whether two concatenated sentences follow each other (or not).

We experiment with two versions of BERT, i.e., BERT\textsubscript{base} and BERT\textsubscript{large}. The former has
12 encoder layers stacked on top of each other, 12 attention heads, and 110M parameters. At each layer, it learns 768-d embeddings. The latter has 24 layers, 16 attention heads, and 340M parameters. It learns 1024-d embeddings. For both models, we use HuggingFace~\cite{wolf2020transformers} implementations.\footnote{\url{https://huggingface.co/bert-base-uncased}\\\url{https://huggingface.co/bert-large-uncased}}

\subsection{Word-Level Representations}\label{sec:representations}

For each triplet in the dataset, we employ BERT models to obtain representations for the compound, the left constituent, and the right constituent. We henceforth use the general term \emph{word} to refer to any of the items in a triplet. We obtain representations for words 
in two conditions, \emph{no-context} (\texttt{NC}) and \emph{in-context} (\texttt{C}), that we describe below.

\paragraph{No-Context (\texttt{NC})}

In this condition, we obtain a single, static representation for a word (e.g., \emph{snowboard}) by feeding it into the model in isolation, i.e., without any surrounding context.
When fed with a single word, BERT outputs embeddings for the tokens which make it up (that result from the tokenization process), as well as for the special tokens [CLS] and [SEP] at the beginning and end of the sequence, respectively. Following previous work~\cite{vulic-etal-2020-probing}, we explore 3 methods for obtaining a word representation. These methods differ with respect to what embeddings are 
taken into account
when building such representation:

\begin{itemize}
    \item \texttt{nospec} This method ignores the special tokens [CLS] and [SEP]. A word representation is built by   averaging the embeddings of the tokens that make up the word (\emph{snow}, \emph{\#\#board}); 
    \item \texttt{withcls} This method builds a word representation by averaging the embedding for the special token [CLS] with the embeddings of the tokens making up the word (\emph{snow}, \emph{\#\#board});
    \item \texttt{all} This method builds a word representation by averaging all the embeddings that are output by BERT for the sequence, i.e., [CLS], [SEP], and the tokens making up the word.
\end{itemize}

\paragraph{In-Context (\texttt{C})}

In this condition, we follow the method by~\citet{bommasani2020interpreting} to obtain a single, static representation of a word from the $N$ contextualized embeddings
produced by BERT for that word in context.
First, we average the
representations of the tokens
that make up a given word---as in the \texttt{NC\_nospec} setting.
Second, we consider all the contextualized representations for a given word   and aggregate them to obtain a single representation that is not dependent on a specific context. We do this by averaging the $N$ contextual representations of a word $w_{1},  \dots, w_{N}$: 
\begin{equation}
w = mean(w_{1},  \dots, w_{N}) \label{eq:static}
\end{equation}

\noindent To obtain contextualized vectors, we sample sentences containing items from our 628 triplets from a cleaned English Wikipedia corpus.\footnote{\url{https://www.lateral.io/resources-blog/the-unknown-perils-of-mining-wikipedia}} For each word, we sample all the sentences in the corpus that contain it, up to a maximum of 100 unique instances per word. The average number of instances per word in our sample is 89.3 (min 1, max 100).

We henceforth simply refer to this setting as \texttt{C}.

\paragraph{Experimental details} Within each setting, we therefore
obtain a single 768-d (BERT\textsubscript{base}) or 1024-d (BERT\textsubscript{large}) embedding for each compound and constituent in our dataset. This operation is performed for each layer of each model---i.e., 12 layers in BERT\textsubscript{base} (1-12) and 24 layers in BERT\textsubscript{large} (1-24). All representations are obtained by running a pre-trained BERT in inference mode, i.e., without fine-tuning or updating the model’s weights.

\paragraph{Baseline} As a baseline, we employ 
static embeddings by GloVe~\cite{pennington2014glove}. We use the 300-d embeddings from the  version of the model trained with 6B tokens.\footnote{\url{https://nlp.stanford.edu/data/glove.6B.zip}}
Since four compounds\footnote{Namely, \emph{livelong}, \emph{sunlamp}, \emph{dunghill}, and \emph{handclasp}.} were not found in GloVe's vocabulary, for this baseline we obtain results for 624 triplets.

\subsection{Predicting Psycholinguistic Measures}

\paragraph{LMD} is a scalar in [0,10] that quantifies the relative semantic role of each constituent toward the meaning of the whole compound:
The higher the value, the more the compound's semantics depends on the right constituent. 
Using BERT's representations for a $\langle$compound, left, right$\rangle$ triplet, we therefore operationalize LMD as follows:
\begin{equation}
{LMD}(c) = 5(R-L)+5
\label{lmd-eq}
\end{equation}
where $c$ is the compound, $L$ is the cosine similarity in [0,1] between the left constituent and the compound, $cos(left,compound)$, and $R$ is the cosine similarity between the right constituent and the compound, $cos(right,compound)$. The scaling and addition operations make the values range in [0,10]. If $L=0$ and $R=1$, then $LMD(c)=10$. \emph{Vice versa}, if $L=1$ and $R=0$, $LMD(c)=0$.

\begin{table}[t!]
	\resizebox{1\columnwidth}{!}{%
\begin{tabular}{llll}
\hline
model      & setting     & \multicolumn{2}{c}{metric (best layer)}   \\ % \hline
           &                    & MAE $\downarrow$            & Spearman $\rho$ $\uparrow$     \\ \hline
GloVe      & --                 & \textbf{0.945} % 0.9452394997395849                   
& 0.541 %0.5406747222818304
            \\ \hline

BERT\textsubscript{base}  
           & \texttt{NC\_nospec}  & 1.095 (11)          & 0.375 (11)          \\
           & \texttt{NC\_all}     & 1.072 (11)          & 0.384 (11)          \\
           & \texttt{NC\_withcls} & 1.071 (11)          & 0.385 (11)          \\
            & \texttt{C}  & 0.991 (11)          & \emph{0.563 (10)}           \\ \hline

BERT\textsubscript{large}
           & \texttt{NC\_nospec}  & 1.130 (21)          & 0.247 (21)           \\
           & \texttt{NC\_all}     & 1.105 (21)          & 0.247 (21)          \\
           & \texttt{NC\_withcls} & 1.107 (22)          & 0.244 (21) \\
           & \texttt{C}  & \emph{0.966 (21)} & \textbf{0.586 (21)} \\ \hline

\end{tabular}
}
\caption{LMD. Results in \textbf{bold} and \emph{italic} are the best and second-best in the column, respectively. Results are from a model's best-performing layer (in parentheses).}\label{tab:resultsLMD}
\end{table}

\paragraph{ST} is a scalar in [1,7] that quantifies the degree to which the meaning of a compound can be inferred from the meaning of the constituents. The higher the value, the more the compound semantics can be inferred from the two constituents' meanings. Using BERT's representations for $\langle$compound, left, right$\rangle$, we operationalize ST as follows:
\begin{equation}
ST(c)=\frac{6(L+R)}{2} + 1
\label{tran-eq}
\end{equation}
where $c$, $L$, and $R$ are defined as above.
The scaling and addition operations make the values range in [1,7]. If $L=1$ and $R=1$, then $ST(c)=7$. \emph{Vice versa}, if $L=0$ and $R=0$, $ST(c)=1$.

\subsection{Evaluation}

We evaluate the effectiveness of each model in approximating human LMD and ST by means of two metrics: mean absolute distance (MAE) and Spearman correlation ($\rho$) between the predicted and human values. For MAE, the lower the distance, the better. For $\rho$, the higher the correlation, the better.

In the next section, we report results by all models in all settings in approximating LMD and ST.

%%%%%%%%%%%%%%%%%%
%%%%%%%%%%%%%%%%%%
%%%%%%%%%%%%%%%%%%
%%%% RESULTS %%%%%
%%%%%%%%%%%%%%%%%%
%%%%%%%%%%%%%%%%%%
%%%%%%%%%%%%%%%%%%

\section{Results}

\subsection{Lexeme Meaning Dominance}\label{sec:resultsLMD}

In Table~\ref{tab:resultsLMD}, we report the results by (the best layer of) each model on LMD in the various settings. Several key observations can be made.
First, both BERT models achieve moderate positive correlation\footnote{As per standard interpretation~\cite{prion2014making}, we consider $\rho$ correlations from ±0.41 to ±0.60 as moderate.} (close to 0.6) with human judgments, with BERT\textsubscript{large} outperforming BERT\textsubscript{base} by some margin.
On the one hand, this indicates that BERT's representations 
do a fairly good job
in accounting for 
the relative semantic weight of each constituent in a 
(lexicalized) compound.
On the other hand, it suggests that more data and parameters play a role in approaching human intuitions.

Second, BERT models outperform GloVe 
in terms of correlation.
This indicates that BERT's embeddings not only encode sensible semantic information~\cite[in line with previous findings; see][]{bommasani2020interpreting,vulic-etal-2020-probing} but also align with 
human semantic intuitions to a greater extent than % the ones obtained with 
the previous-generation GloVe model.
However, it is worth noting that BERT models outperform GloVe only in  \texttt{C}, but not in \texttt{NC}. This clearly shows that BERT's embeddings have an advantage over GloVe's ones only when leveraging information in the surrounding context and reveals that BERT struggles to represent out-of-context words, likely due to its architecture and training regime. In~\ref{sec:templates}, we report that the lack of \textit{any} surrounding context in \texttt{NC} is indeed detrimental to model representations, though sensible contextual information in \texttt{C} is needed to properly approximate LMD.
Moreover, GloVe achieves the lowest MAE, which 
shows that these embeddings are effective in approximating the raw LMD values---though contextualized \texttt{C} BERT representations are better at capturing the overall pattern of similarity.\footnote{Recall that MAE measures the distance between the target and predicted values, while Spearman $\rho$ quantifies strength and direction of association between the two ranked variables.}

\begin{figure}[t!]
	\centering
	\includegraphics[width=1\linewidth]{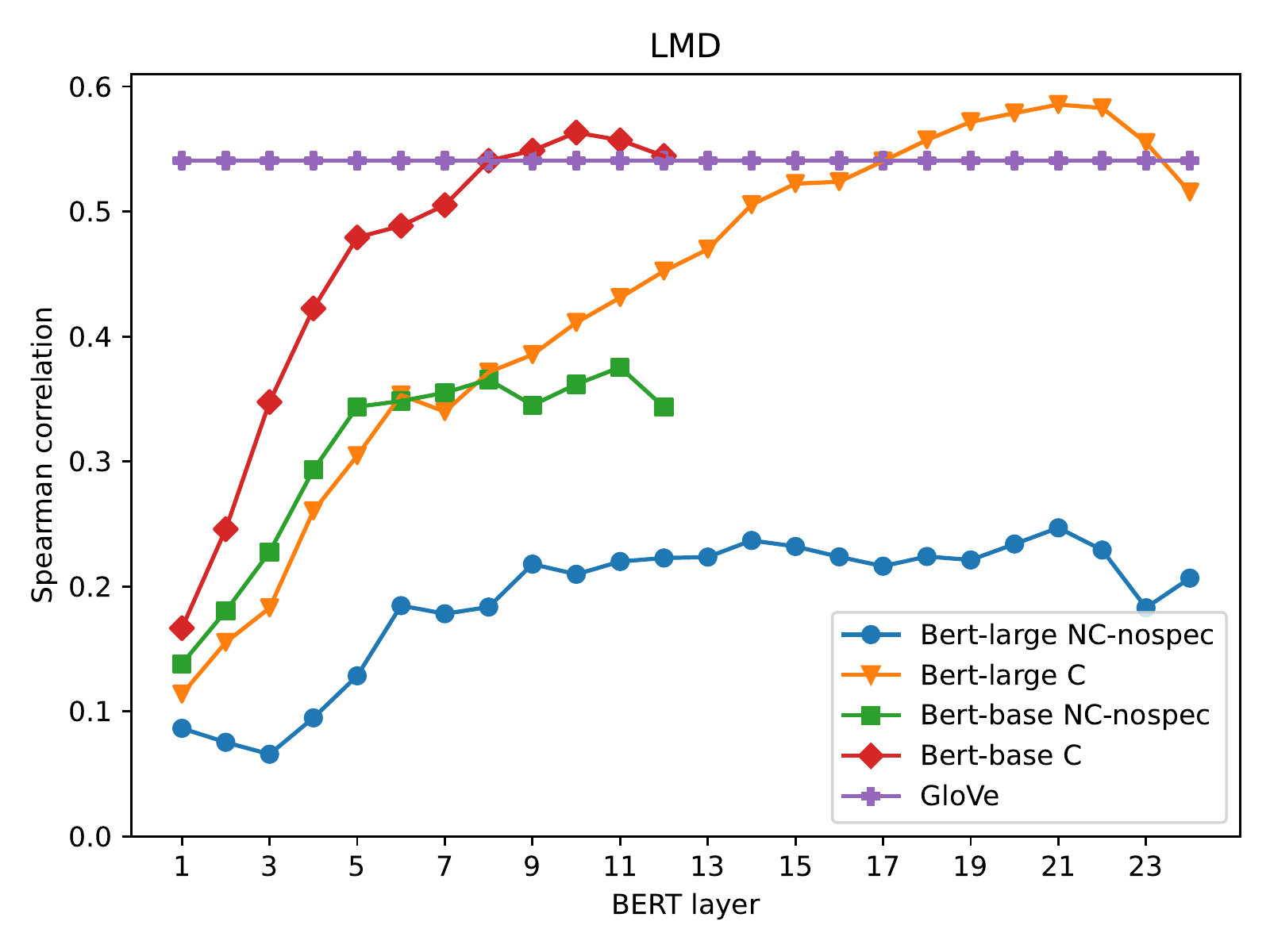}
	\caption{LMD. $\rho$ against model layers. For out-of-context BERT models, we only report the best-performing \texttt{nospec} setting. Best viewed in color.}
	\label{fig:plotLMD}
\end{figure}

Third, the pattern of correlation values over the model's layers highlights the importance of contextualization for compound representation.
As can be seen in Figure~\ref{fig:plotLMD}, best-performing \texttt{C} BERT models show 
an almost constant increasing trend,
with the highest correlation being achieved in high layers---layer 9 and 20 in BERT\textsubscript{base} and BERT\textsubscript{large}, respectively.
This is an opposite pattern compared to what was observed in previous work, where lower layers were found to 
encode
most lexical semantic information~\cite{bommasani2020interpreting,vulic-etal-2020-probing}. These patterns do not necessarily contradict each other. Indeed, we argue that judging  the semantic relationship between a compound (e.g., \textit{handgun}) and its constituent words (\textit{hand}, \textit{gun}) might involve relying on a specific
interpretation of the compound
rather than on abstract
lexico-semantic information. Since previous work showed that 
word senses are better encoded in deeper layers~\cite{reif2019visualizing},
this could explain why  these layers are also good at capturing LMD.
A similar---though flatter---trend is observed for \texttt{NC} models.

% Good:
% ponytail, high (), dist 0.196; wartime, low (3.47), dist 0.054\\

% Bad:
% milestone, low (3.36), dist 1.272; muskrat, high (7.53), dist 2.846\\

\begin{table}[]
	\resizebox{1\columnwidth}{!}{%
\begin{tabular}{llll}
\hline
model  & setting   & \multicolumn{2}{c}{metric (best layer)}   \\
       &           & MAE $\downarrow$            & Spearman $\rho$ $\uparrow$     \\ \hline
GloVe  & --        & 2.657 %  2.657143596740566         
& 0.304 % 0.30420692154328727              
\\ \hline
BERT\textsubscript{base} & \texttt{NC\_nospec}  & 0.953 (6)           & 0.316 (5)           \\
                         & \texttt{NC\_all}     & 1.129 (10)          & 0.234 (1)           \\
                         & \texttt{NC\_withcls} & 0.989 (1)           & 0.275 (3)           \\
                         & \texttt{C}  & \emph{0.899 (9)}    & \emph{0.415 (9)}    \\ \hline
BERT\textsubscript{large}& \texttt{NC\_nospec}  & 0.989 (9)           & 0.195 (6)           \\
                         & \texttt{NC\_all}     & 1.118 (24)          & 0.113 (1)           \\
                         & \texttt{NC\_withcls} & 1.024 (6)           & 0.139 (1)           \\
                         & \texttt{C}  & \textbf{0.876 (19)} & \textbf{0.476 (20)} \\ \hline
\end{tabular}
}
\caption{ST. Results in \textbf{bold} and \emph{italic} are the best and second-best in the column, respectively. Results are from a model's best-performing layer (in parentheses).}\label{tab:resultsST}
\end{table}

\begin{figure}[b!]
	\centering
	\includegraphics[width=1\linewidth]{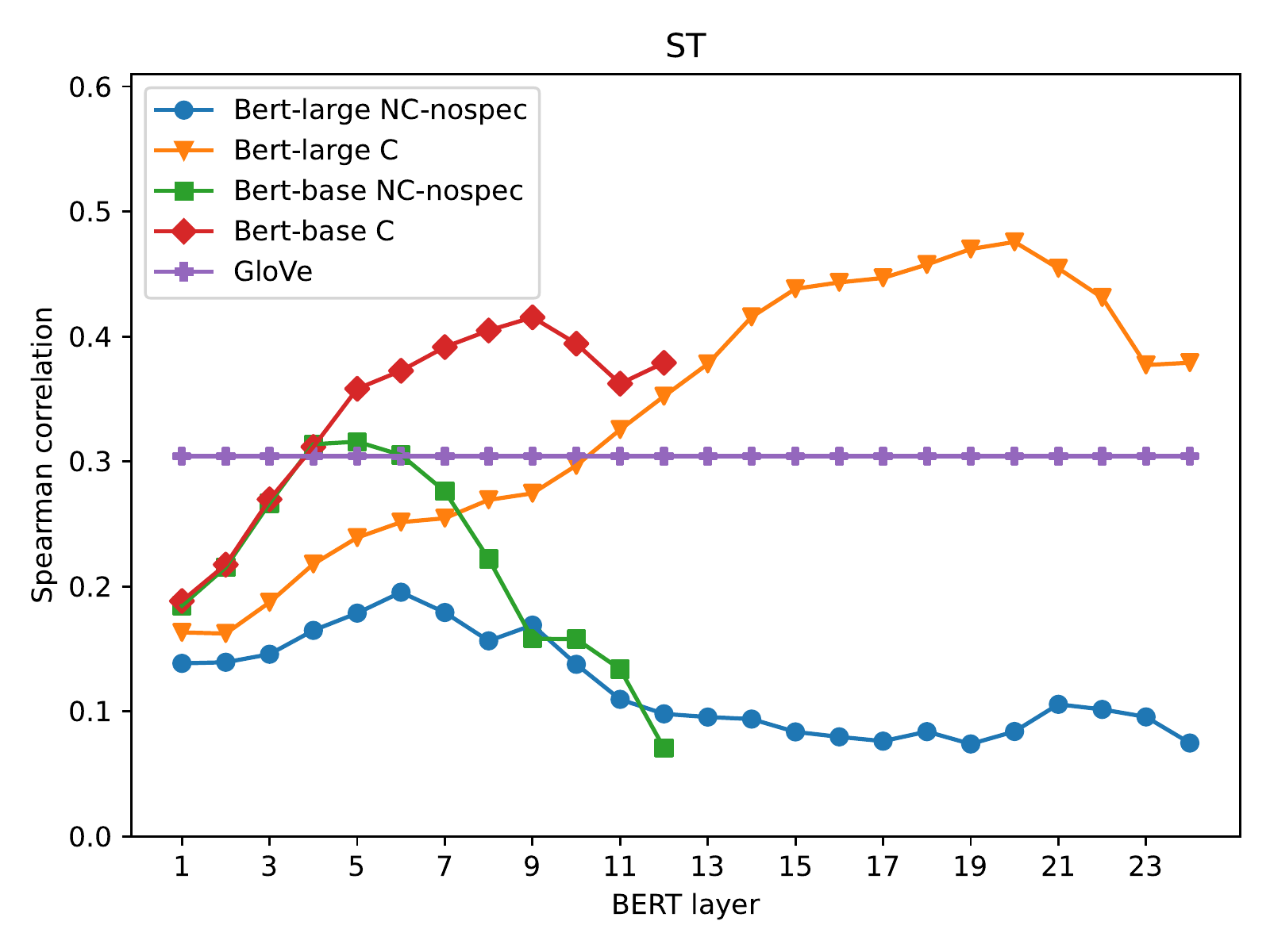}
	\caption{ST. $\rho$ against model layers. For out-of-context BERT models, we only report the best-performing \texttt{nospec} setting. Best viewed in color.}
	\label{fig:plotST}
\end{figure}

\subsection{Semantic Transparency}\label{sec:resultsST}

In Table~\ref{tab:resultsST}, we report the results by (the best layer of) each model on ST in the various settings. Several key observations can be made. First, both BERT models achieve moderate positive correlation with human judgments, with BERT\textsubscript{large} outperforming BERT\textsubscript{base}. This indicates that BERT's representations are moderately effective in predicting the extent to which a compound meaning is recoverable from the meaning of its constituents
and that more data and parameters help---indeed, the gap between the two BERT models is higher here than in LMD (0.06 vs 0.02). At the same time, the highest correlation achieved by BERT\textsubscript{large} (0.476) on ST is substantially lower than on LMD (0.586), which indicates that 
ST is more challenging to approximate compared to LMD.

Second, BERT models outperform GloVe on both metrics. Though correlations are generally lower than in LMD, the gap between BERT models and GloVe is much more pronounced here. That is, BERT's contextualized embeddings have an even clearer advantage over previous-generation ones in modeling ST compared to LMD.

Third, as can be seen in Figure~\ref{fig:plotST}, the overall best results are achieved by \texttt{C} embeddings in high layers---layer 8 and 19 for BERT\textsubscript{base} and BERT\textsubscript{large}, respectively---which replicates the findings for LMD. This confirms the role of context and 
contextualization
for obtaining better representations of compounds.\footnote{See the analysis in~\ref{sec:templates} for further evidence of the role of sensible semantic context in approximating ST.}
Interestingly, in \texttt{NC} settings, BERT models show a 
different pattern compared to LMD, with correlation reaching a `peak' within the first layers and then constantly
decreasing.
This suggests that decontextualized lexico-semantic information encoded in lower layers accounts for ST to some extent, on par with or even outperforming GloVe (this is the case for BERT\textsubscript{base} \texttt{NC\_nospec}).

\begin{figure}
	\centering
	\begin{subfigure}[b]{0.49\columnwidth}
		\centering
  	\caption{ponytail}
		\includegraphics[width=\textwidth]{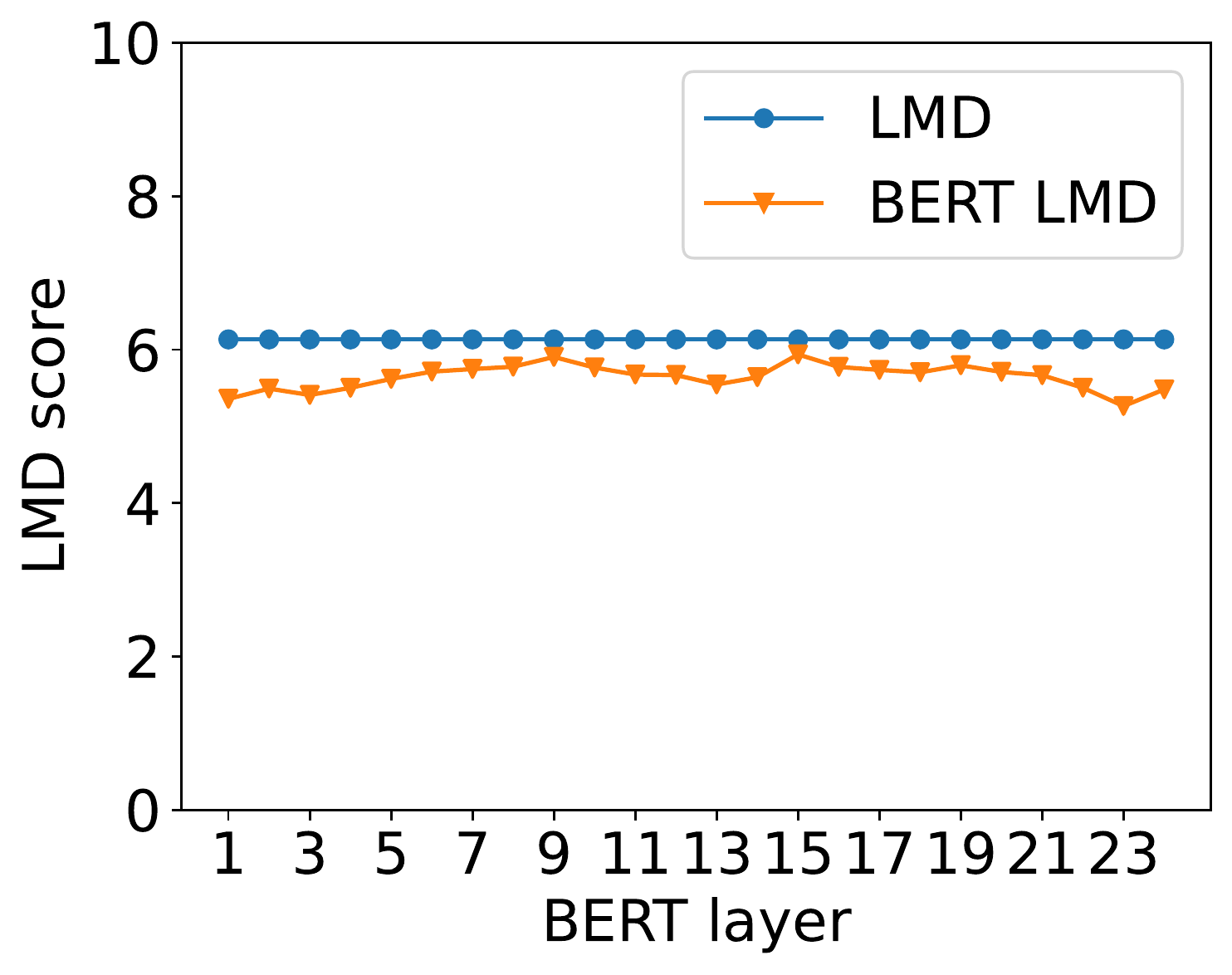}
	\end{subfigure}
	%\hfill
	% \hspace{0.10\textwidth}
    \centering
	\begin{subfigure}[b]{0.49\columnwidth}
		\centering
  	\caption{wartime}
		\includegraphics[width=\textwidth]{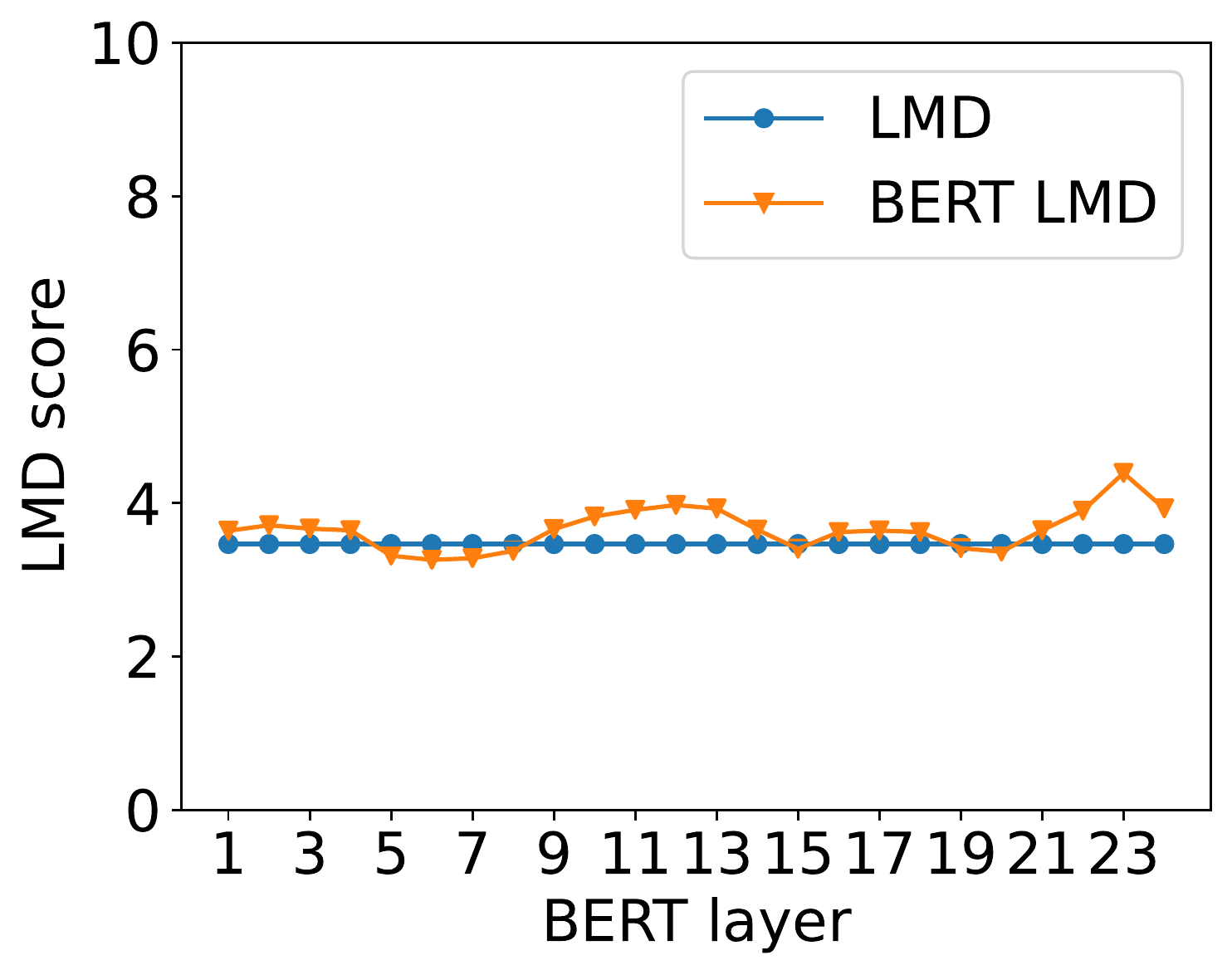}
	\end{subfigure}

	\centering
	\begin{subfigure}[b]{0.49\columnwidth}
		\centering
  	\caption{muskrat}
		\includegraphics[width=\textwidth]{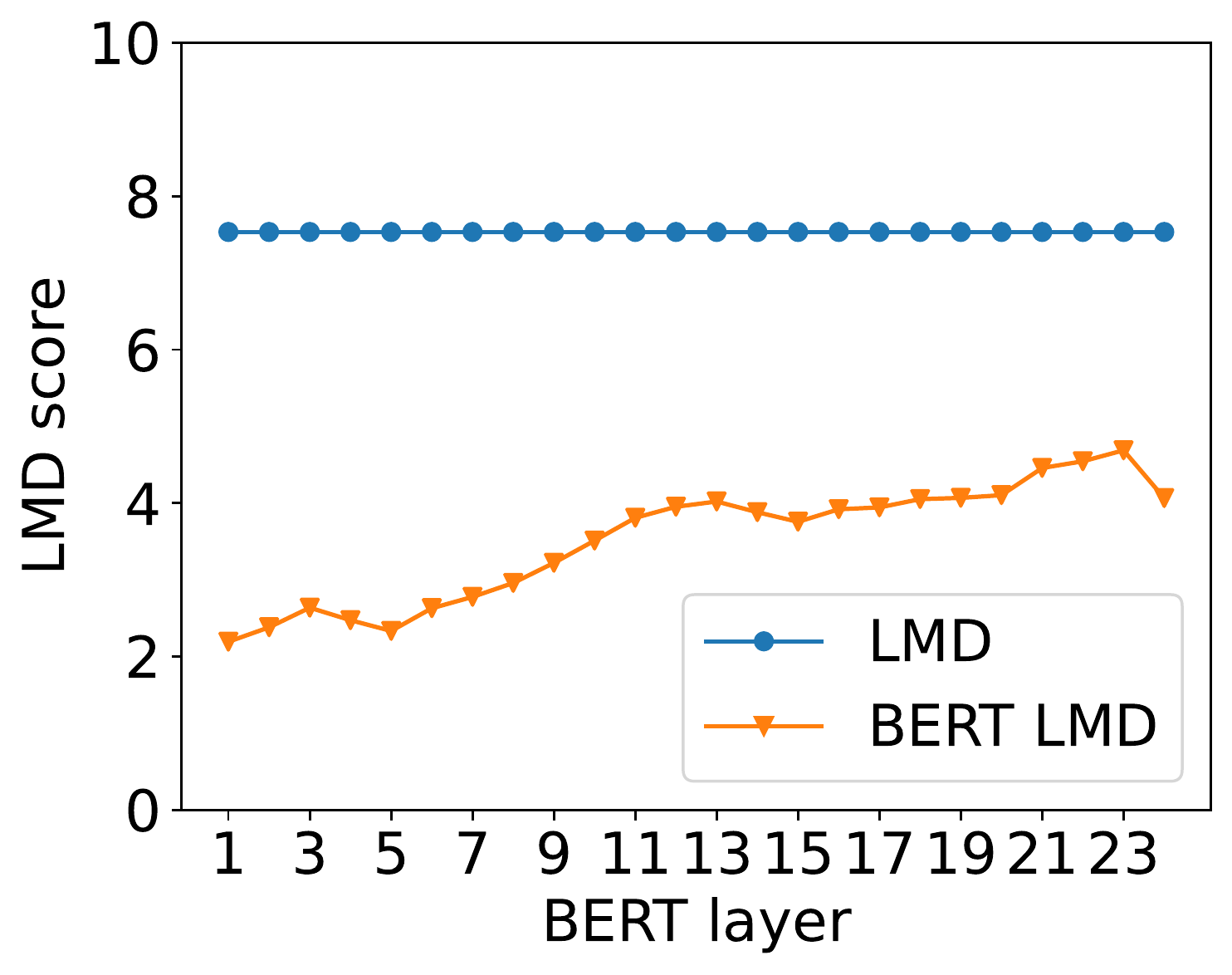}
	\end{subfigure}
	%\hfill
	% \hspace{0.10\textwidth}
    \centering
	\begin{subfigure}[b]{0.49\columnwidth}
		\centering
  	\caption{milestone}
		\includegraphics[width=\textwidth]{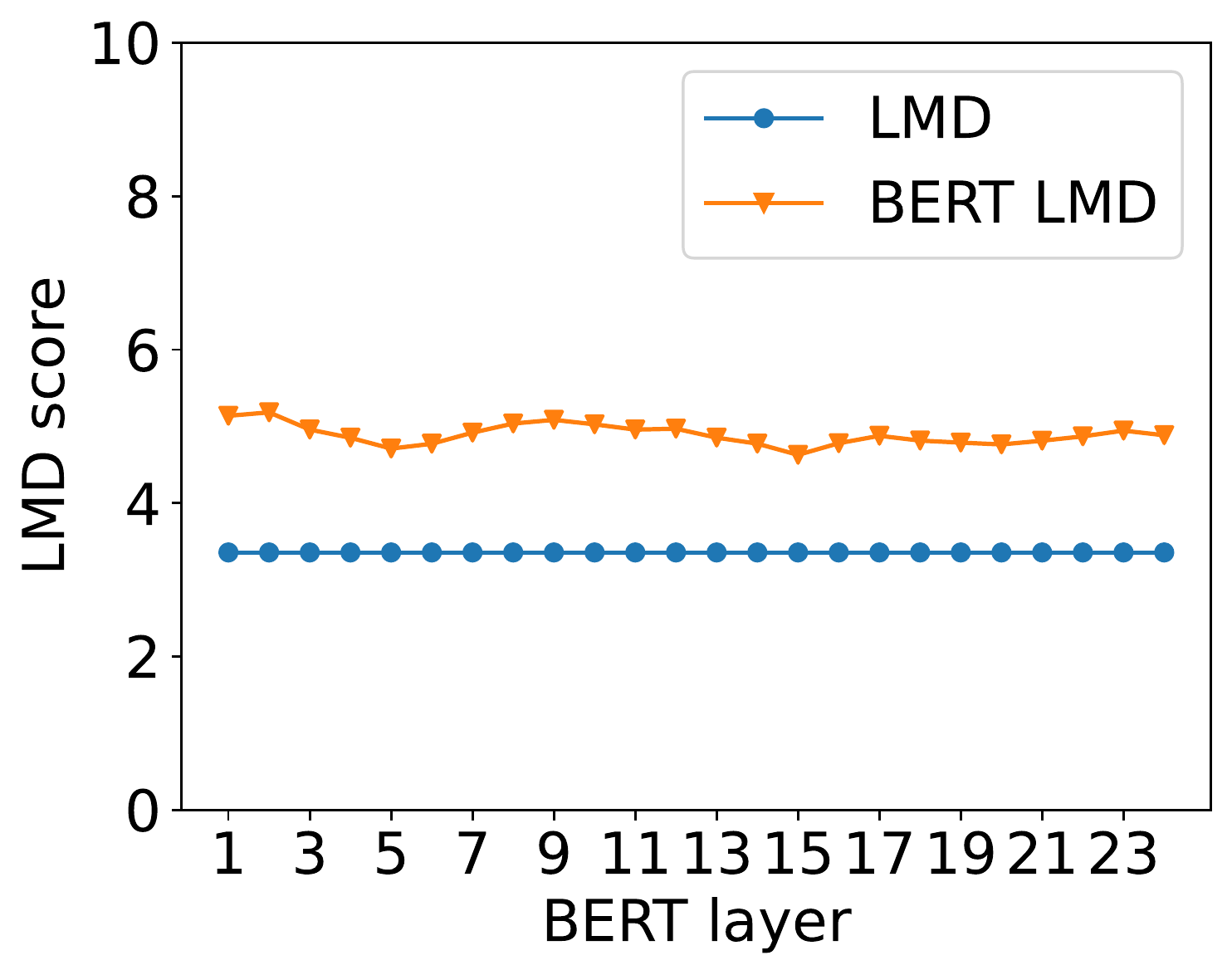}
	\end{subfigure}
	\caption{Examples where \texttt{C} BERT\textsubscript{large} is good (top) and bad (bottom) in approximating human LMD.
	From top left, clockwise: \emph{ponytail}, \emph{wartime}, \emph{milestone}, \emph{muskrat}.}

	\label{fig:LMDcases}
\end{figure}

\subsection{Examples}

In Figure~\ref{fig:LMDcases} 
we report some examples where the best-performing \texttt{C} BERT\textsubscript{large} is good (top) and bad (bottom) in approximating LMD. As can be seen, the distance between the predicted and human LMD is extremely low for \emph{ponytail} 
% (0.196 distance) 
and \emph{wartime},
% (0.054), 
namely, a high-LMD (6.13) and a low-LMD (3.47) compound, respectively. In contrast, the distance is very large for high-LMD (7.53)
\emph{muskrat}
% (2.846 distance) 
and low-LMD (3.36) \emph{milestone}. While the predicted LMD values are fairly stable over the layers for \textit{ponytail}, \textit{wartime}, and \textit{milestone}, for \textit{muskrat} the higher layers are better to approximate the real LMD by assigning an increasingly higher semantic weight to the \textit{rat} lexeme. This could be due to the representation of \textit{muskrat} becoming more `aware'---through contextualization---of the  semantic traits related to the animal domain, apparently less present in earlier layers. Also, it is interesting to note that for the compound \textit{milestone}, which is `exocentric' (i.e., the head is neither \textit{mile} nor \textit{stone}), BERT keeps predicting a conservative LMD value, which similarly weights the two constituents, over the layers. That is, contextualization does not make \textit{mile} or \textit{stone} become dominant in the compound (while, interestingly, human speakers consider \textit{mile} as slightly dominant over \textit{stone}).
% (1.272).

\begin{figure}
	\centering
	\begin{subfigure}[b]{0.49\columnwidth}
		\centering
		\caption{policeman}
		\includegraphics[width=\textwidth]{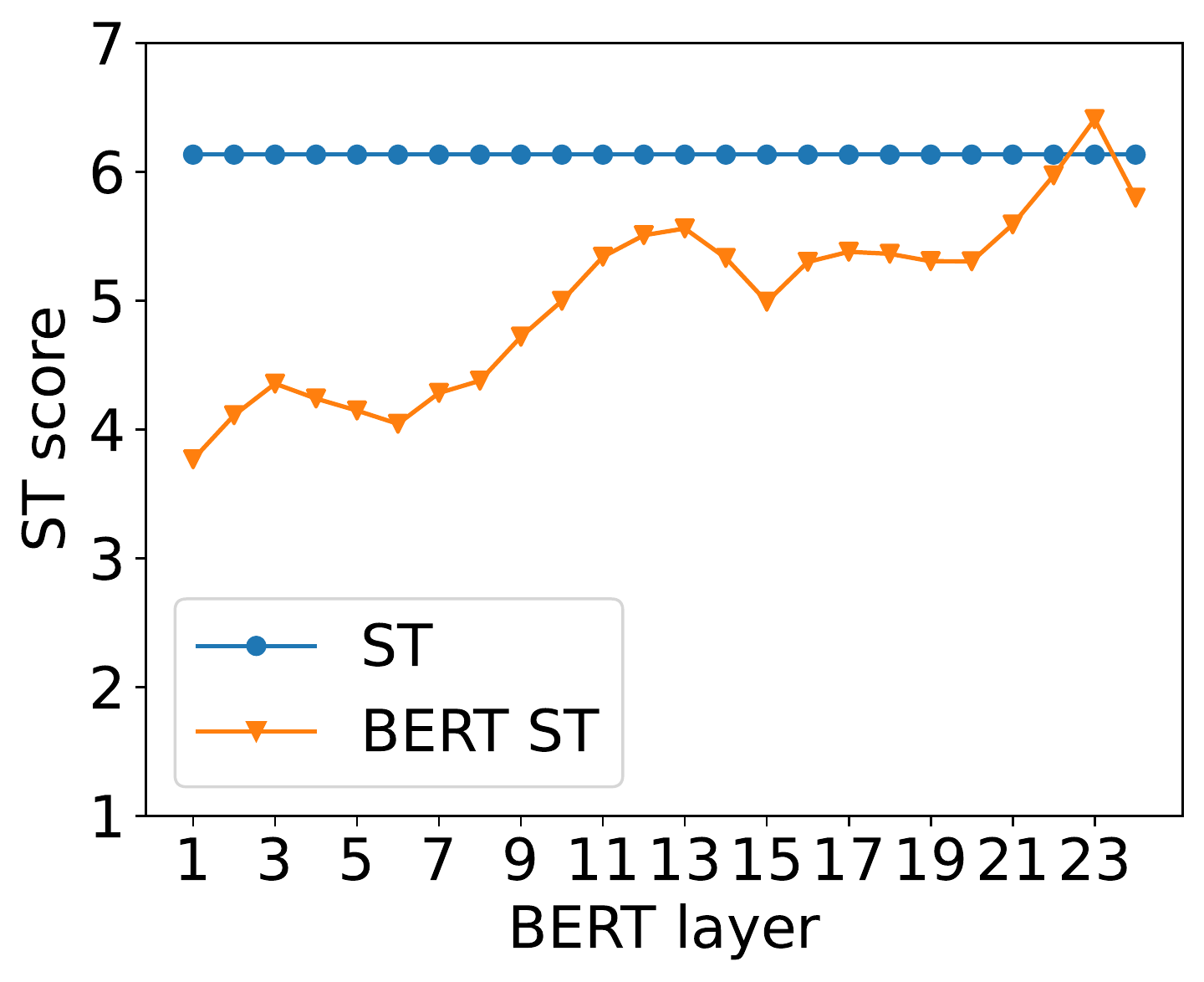}
	\end{subfigure}
	%\hfill
	% \hspace{0.10\textwidth}
    \centering
	\begin{subfigure}[b]{0.49\columnwidth}
		\centering
		\caption{milestone}  
		\includegraphics[width=\textwidth]{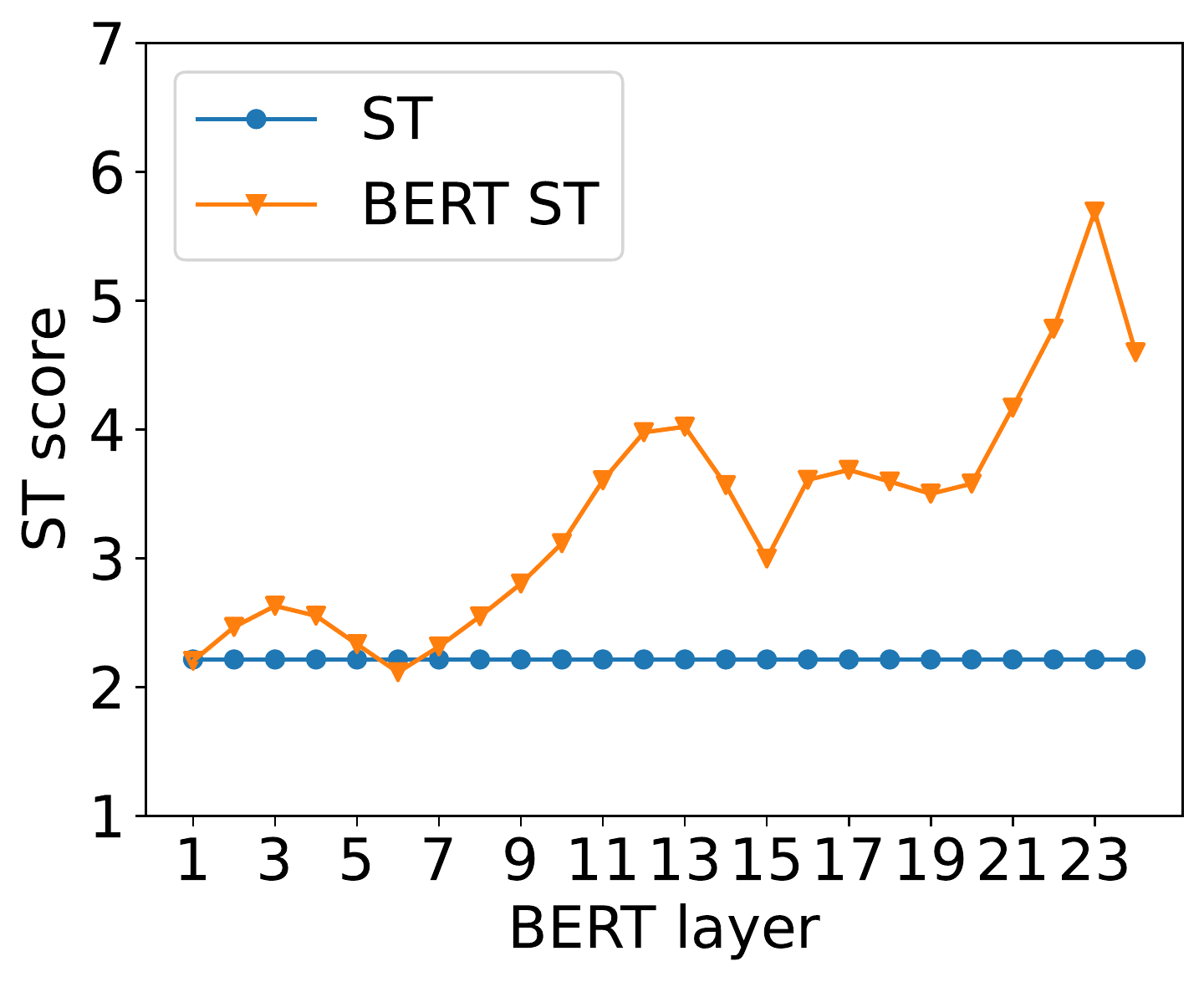}
	\end{subfigure}

	\centering
	\begin{subfigure}[b]{0.49\columnwidth}
		\centering
		\caption{muskrat}
		\includegraphics[width=\textwidth]{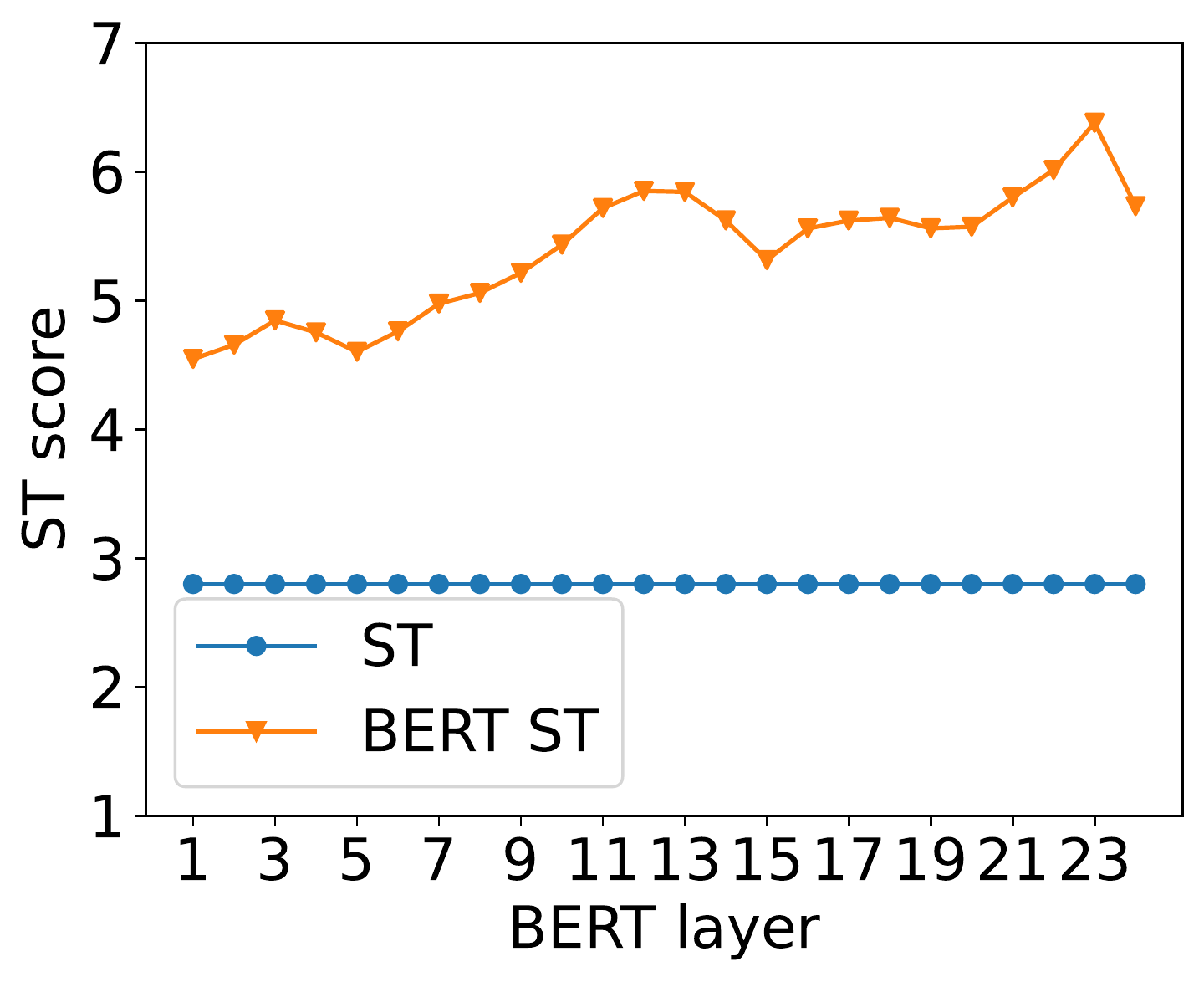}
	\end{subfigure}
	%\hfill
	% \hspace{0.10\textwidth}
    \centering
	\begin{subfigure}[b]{0.49\columnwidth}
		\centering
		\caption{cheapskate}
		\includegraphics[width=\textwidth]{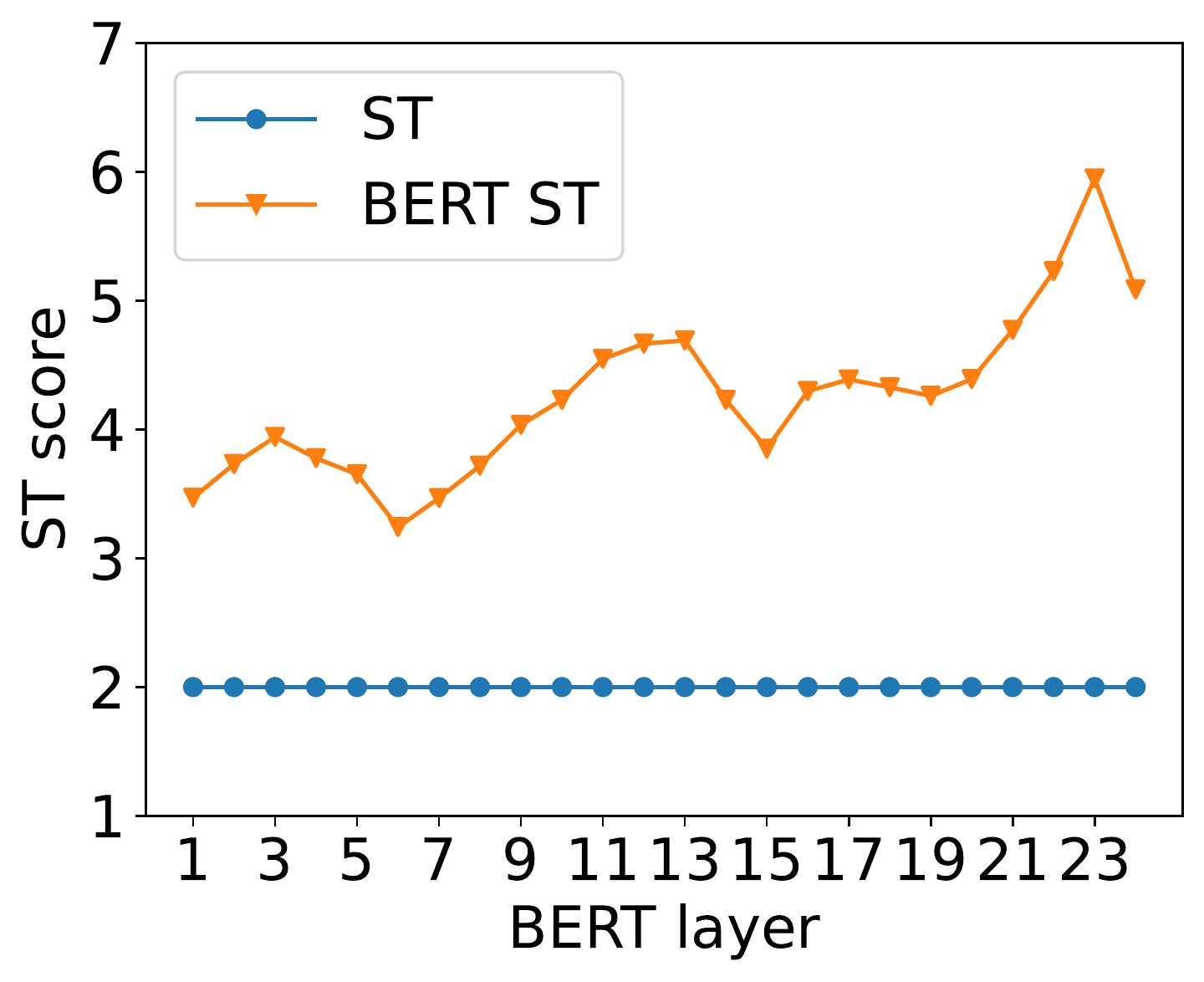}
	\end{subfigure}
	\caption{Examples where \texttt{C} BERT\textsubscript{large} is good (top) and bad (bottom) on ST. From top left, clockwise: \emph{policeman}, \emph{milestone}, \emph{cheapskate}, \emph{muskrat}.}
	\label{fig:STcases}
\end{figure}

In Figure~\ref{fig:STcases}, we report
some good (top) and bad (bottom) cases for the same model in approximating ST.
As can be seen, the distance between the predicted and human ST is very low for high-ST (6.31) \emph{policeman}
% (0.161 distance) 
and low-ST (2.21) \emph{milestone}.
% (0.012). 
In contrast, the distance is large for low-ST (2.80) 
\emph{muskrat} 
% (1.746 distance) 
and low-ST (2.00) \emph{cheapskate}.
% (1.241).
Moreover, it can be noted that higher layers are better than lower layers in approximating ST for \textit{policeman}, which is highly transparent, while they are worse for \textit{milestone}, \textit{cheapskate}, and \textit{muskrat}, which are very opaque. This goes hand in hand with a seemingly general trend observed in these examples: the more contextualization, the higher the ST. This could reflect a generalized increase of cosine similarity values through BERT's layers~\cite[as reported by][]{ethayarajh-2019-contextual}, which would lead in turn to a higher ST by virtue of how ST was operationalized; see Eq.~\ref{tran-eq}. However, our results (see Figure~\ref{fig:plotST}) show that this increase does not correspond to a higher correlation with human judgments: indeed, correlation steadily decreases in the last 4 layers of BERT\textsubscript{large}, where cosine similarities are highest.

Finally, while BERT struggles on both LMD and ST for \emph{muskrat}, for \emph{milestone} it struggles on LMD but does a good job on ST. This confirms that LMD and ST capture different semantic aspects: a good performance on one  measure does not necessarily guarantee a good performance on the other.\footnote{Computing the correlation between LMD and ST human values in the dataset (-0.013) confirms this intuition.}

% This is confirmed by computing the correlation of human LMD and ST values in the dataset (-0.013).}
% and non-significant ($p$=0.749).}

\subsection{Which Factors Drive the Prediction?}\label{sec:R}

To more formally investigate which factors contribute to higher predicted values of LMD and ST by the best performing model \texttt{C} BERT\textsubscript{large}, we run two linear regression models in \texttt{R}---one for LMD, one for ST---using, for each compound, the predicted LMD/ST value by the best layers (21/20, respectively) as the dependent variable, and the following independent variables: (1) the number of tokens into which the compound was split by the tokenizer, e.g., 2 for \textit{snowboard} (\textit{snow}, \textit{\#\#board}); (2) the frequency of the compound in our dataset, i.e., the number of instances on the top of which the average representation was computed; (3) the compound concreteness; (4) the modifier (left constituent) concreteness; (5) the head (right constituent) concreteness. Concreteness values are extracted from~\citet{brysbaert2014concreteness}.\footnote{23 compounds out of 628 were not present in the concreteness database and therefore excluded from the analysis.}

For LMD, both the concreteness of the head and the modifier---but not other variables---have a statistically significant role, though in the opposite direction: the higher the former, the higher the LMD (i.e., more weight to the head); the higher the latter, the lower the LMD (i.e., more weight to the modifier). This makes intuitive sense and shows that BERT assigns more `weight' to concrete constituents.
For ST, three variables have a statistically significant role in predicting higher values, and all in the same direction: the higher the number of tokens, the compound concreteness, and the modifier concreteness, the higher the ST. As for concreteness, this generally shows that BERT assigns higher similarities to concrete words. As for the effect of the number of tokens, this is an interesting finding, which reveals that BERT considers as more transparent those compounds than can be routinely broken into parts.
The full tables reporting all the effects and corresponding coefficients and p-values can be found in~\ref{sec:statistics}.

\section{Analysis}

\subsection{LMD: Reversed Compounds}

From the results reported in section~\ref{sec:resultsLMD}, it appears that BERT is capable of obtaining sensible representations of compounds that encode the relation between the constituents and their respective semantic `weight'. However, it might still be that the reported moderate correlations result from the model assigning a default high/low similarity to the constituents while being no or little aware of the semantic and syntactic (i.e., the modifier/head) relation which ties them. If that is the case, the model would consider, e.g., the contribution of \emph{war} in \emph{wartime} and \emph{timewar} to be identical---though the meaning of the \emph{reversed} compound would be intuitively very different. As such, we might expect a similar/same LMD value assigned by the model to these two compounds, and therefore a similar correlation with human judgments. Otherwise, if BERT represents a compound by genuinely accounting for the relationship between its constituents, the predicted LMD for the reversed compound (\emph{timewar}) is likely to be different. As such, we might expect a much lower correlation with human intuitions.

\begin{figure}[t!]
	\centering
	\includegraphics[width=1\linewidth]{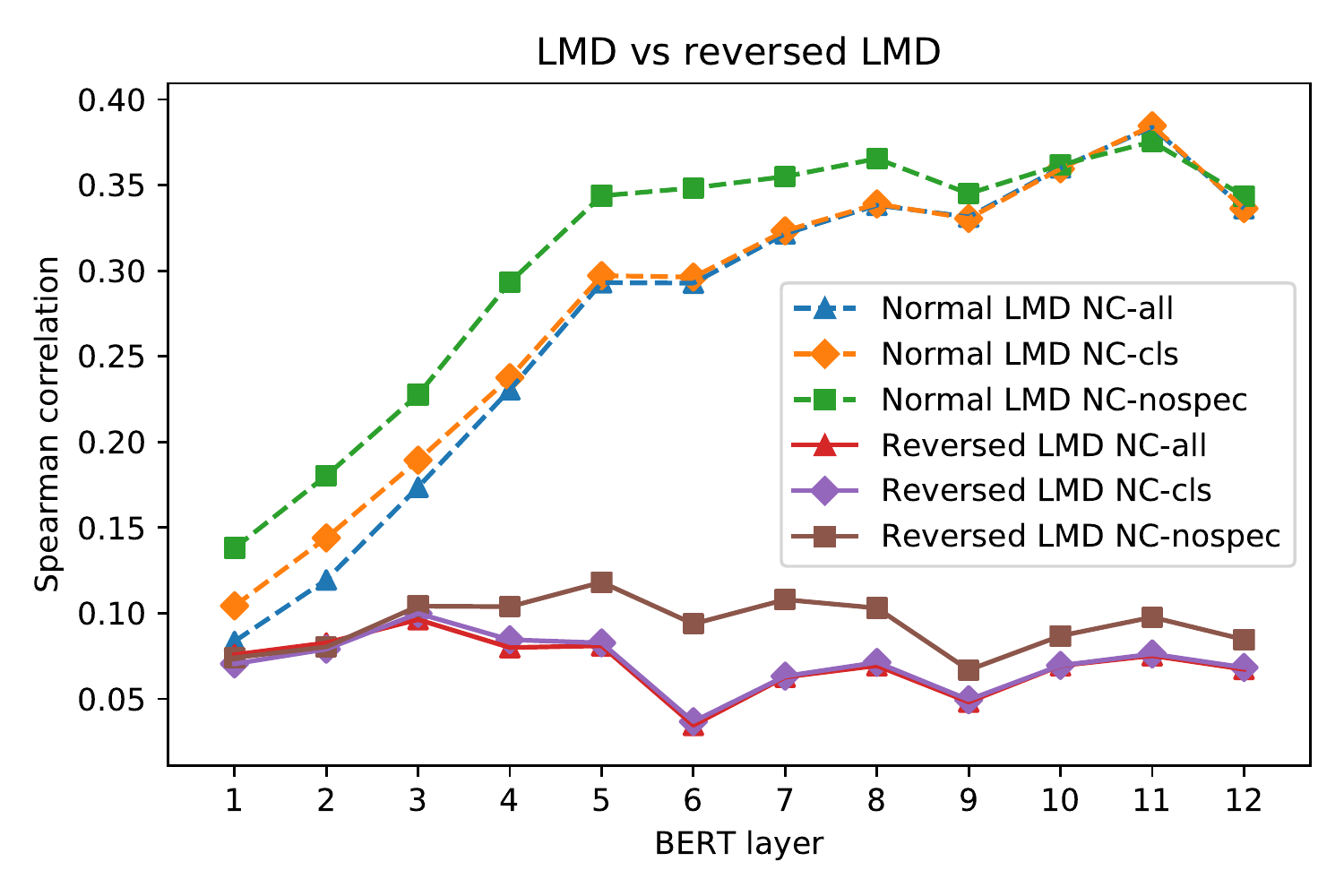} % \hfill
	\caption{$\rho$ for LMD vs LMD \emph{reversed} values  by out-of-context BERT\textsubscript{base} across layers. Best viewed in color.}
	\label{fig:revLMD}
\end{figure}

In this analysis, we test this issue by
re-running the LMD experiment on the \emph{reversed} version of the compounds in our dataset, i.e., \emph{wartime} > \emph{timewar}, \emph{bodyguard} > \emph{guardbody}, etc.
LMD is computed exactly as above, except that we replace the representation of the compound with that of its reversed version.
Since (most of) these reversed compounds are unlikely to occur in standard \emph{corpora} of texts, we experiment with \texttt{NC} representations. Moreover, we
experiment with 
BERT\textsubscript{base} since it outperformed BERT\textsubscript{large} on this setting. 
As can be seen in Figure~\ref{fig:revLMD},
% bertbase_nocontext_all_reversedlmd 0.09630926105093227
% bertbase_nocontext_withcls_reversedlmd 0.09989219026921677
% bertbase_nocontext_nospec_reversedlmd 0.11811757727997432
%%%%%%%%%%%%
the correlation for reversed compounds is much lower than the original one (0.11
vs 0.39).
This is a good sign, which confirms that BERT
does not rely on shortcuts when representing compound meanings and the relative weight of each constituent. Instead, it appears that the model can account for the semantic and syntactic relation between the constituents, even when information from the surrounding context is not available.

\subsection{ST: Weighted Constituents}

From the results in section~\ref{sec:resultsST}, it appears that BERT can approximate, to some extent, the degree to which the meaning of a compound is recoverable from the semantics of the constituents.
When operationalizing ST we assumed that the two constituents are equally responsible for the overall ST---i.e., we computed the unweighted average between the two pairwise similarities.
This way, a high (low) similarity between the compound and one of its constituents would not determine a high (low) ST on its own.
This operationalization is in line with psycholinguistic literature, according to which we have a fully transparent compound when both lexemes contribute to its meaning and a fully opaque compound when neither of the two contribute~\cite[see, e.g.,][]{libben1998semantic}.
However, it is an open question whether BERT's compound representations do encode both constituents equally, or whether they disproportionately encode one constituent over the other. If the latter is the case, weighing one 
more than the other
when computing ST
would possibly result in a higher correlation with human judgments. If both constituents are equally represented in the compound embedding, instead, the 
unweighted version---our main experiment---would lead to the highest correlation.

\begin{figure}[t!]
	\centering
	\includegraphics[width=1\linewidth]{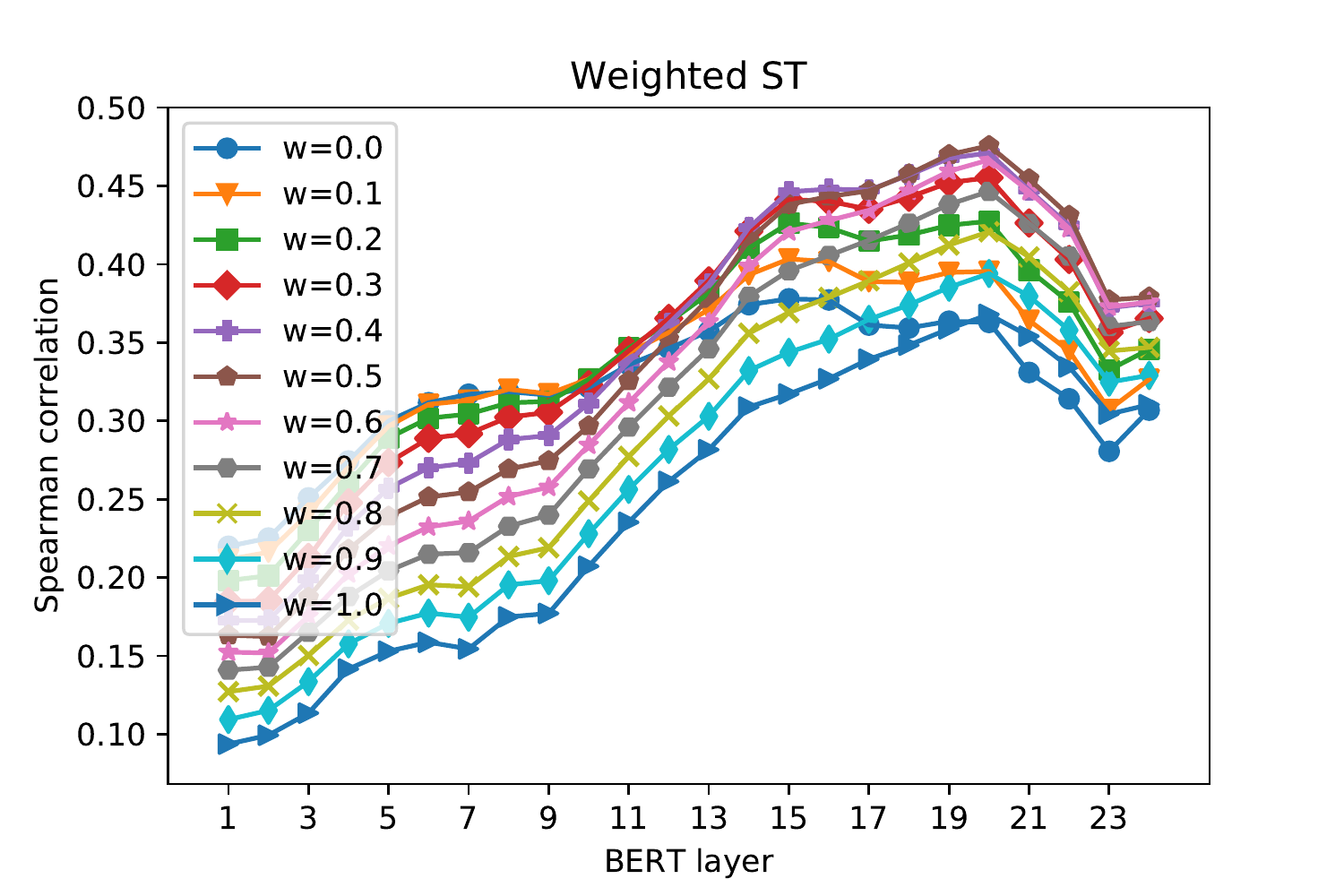} % \hfill
	\caption{$\rho$ for \emph{weighted} ST by in-context BERT\textsubscript{large} across layers. Each weight stands for the weight assigned to the left constituent. Best viewed in color.}
	\label{fig:weightST}
\end{figure}

In this analysis, we test this issue by computing a \emph{weighted} version of ST where the left and right constituents are assigned different weights. For example, we assign weights 0.0 and 1.0 to the left and right constituent, respectively (in this case, only the right one, the compound's  \textit{head}, but not the left one, the \textit{modifier}, will be responsible for ST); 0.1 and 0.9; 0.2 and 0.8; and so on. We experiment with all combinations of weights (including 0.5 and 0.5), which sums up to 11 versions of \emph{weighted} ST.

Figure~\ref{fig:weightST} reports the results of this analysis for the best-performing \texttt{C} BERT\textsubscript{large} model---note that the weight refers to the weight assigned to the left constituent. As can be seen, the highest correlation is achieved by the unweighted ST (weight 0.5) at layer 20. Since these are the results reported in Table~\ref{tab:resultsST}, 
this finding  confirms that both constituents are equally accounted for in the BERT's compound representation. Interestingly, weights that are close to 0.5 perform reasonably well (0.4 and 0.6 rank second and third, respectively), while correlation decreases the more we 
move away from this value. 

Overall, the results of these two analyses confirm that 
BERT  accounts for the left and right constituents equally when representing the semantics of a compound.
Moreover, these representations seem to encode the complex semantic and syntactic relationship tying the compound constituents.

\section{Conclusion}

In this paper, we study how BERT represents the meaning of lexicalized compounds. We take a psycholinguistic angle and show that the model does a reasonably good job of making semantic judgments in line with those of human speakers. Since higher, more contextualized layers are shown to correlate best with human intuitions, we propose that speakers may access a specific, context-dependent representation when making these judgments. In future work,
recent approaches to multimodal word representation in Transformers~\cite{pezzelle-etal-2021-word} could be leveraged to test the role of visual grounding in compound semantics~\cite{gunther2020semantic}.

\section*{Limitations}

\paragraph{Impact of the corpus} To build contextualized representations for compounds, we sample sentences from a corpus of texts. A limitation of our work lies in the use of encyclopedic data only, which limits the number and variety of contextualized meanings a compound can have. Further attention should be paid to this aspect.

\paragraph{Operationalization of the measures} While defining LMD and ST based on the cosine similarity between a compound and its constituent makes intuitive sense, it may not be the only (nor the best) way to operationalize the two measures. Further
exploration on how to formally define them based on the model embeddings should be carried out.

\section*{Ethics Statement}

\paragraph{Broader impact} We do not see any serious ethical problem connected to this research. At the same time, we are aware of the risks associated with the development and use of large NLP models that we use in this research. Such risks include the environmental impact of the computational resources required for training and the encoding and possible amplification of biases present in the massive amounts of un-curated data the models learn from.

% Scientific work published at EACL 2023 must comply with the \href{https://www.aclweb.org/portal/content/acl-code-ethics}{ACL Ethics Policy}. We encourage all authors to include an explicit ethics statement on the broader impact of the work, or other ethical considerations after the conclusion but before the references. The ethics statement will not count toward the page limit (8 pages for long, 4 pages for short papers).

\section*{Acknowledgements}

We would like to thank Marco Marelli for the idea of testing different weights when computing ST. We are grateful to the anonymous EACL reviewers and meta-reviewer for the insightful feedback.

\bibliography{anthology,custom}
\bibliographystyle{acl_natbib}

\appendix

\section{Appendix}

\subsection{Templated Linguistic Contexts}\label{sec:templates}

In this analysis, we investigate the extent to which the disadvantage of \texttt{NC} compared to \texttt{C} in approximating LMD and ST is due to the lack of \textit{any} linguistic context surrounding the compound and the constituent words.\footnote{We thank an anonymous reviewer for suggesting this analysis.} 
Since BERT has probably seen very few examples of words out of context during training, it could be that the model is poor at handling words in isolation, which would have an impact on the resulting representations---and LMD/ST values.
To test this issue, we consider the best-performing BERT\textsubscript{large} and use it to obtain a single, contextualized representation for words (either compounds or constituents) by embedding them in the following templated sentence: \textit{This is a} \texttt{<word>}.
This setting bears similarities with both \texttt{NC} and \texttt{C}. On the one hand, we compute a single representation for each word, similarly to \texttt{NC}. On the other hand, the representation of each word is contextualized (i.e., embedded in a linguistic context), though the surrounding context does not contain any meaningful semantic information.  
As such, we expect these representations to be better than \texttt{NC} by virtue of their higher similarity with standard training samples, but worse than \texttt{C} since they lack any sensible semantic information coming from the context surrounding the word at inference time.

\begin{figure}[t!]
	\centering
	\includegraphics[width=1\linewidth]{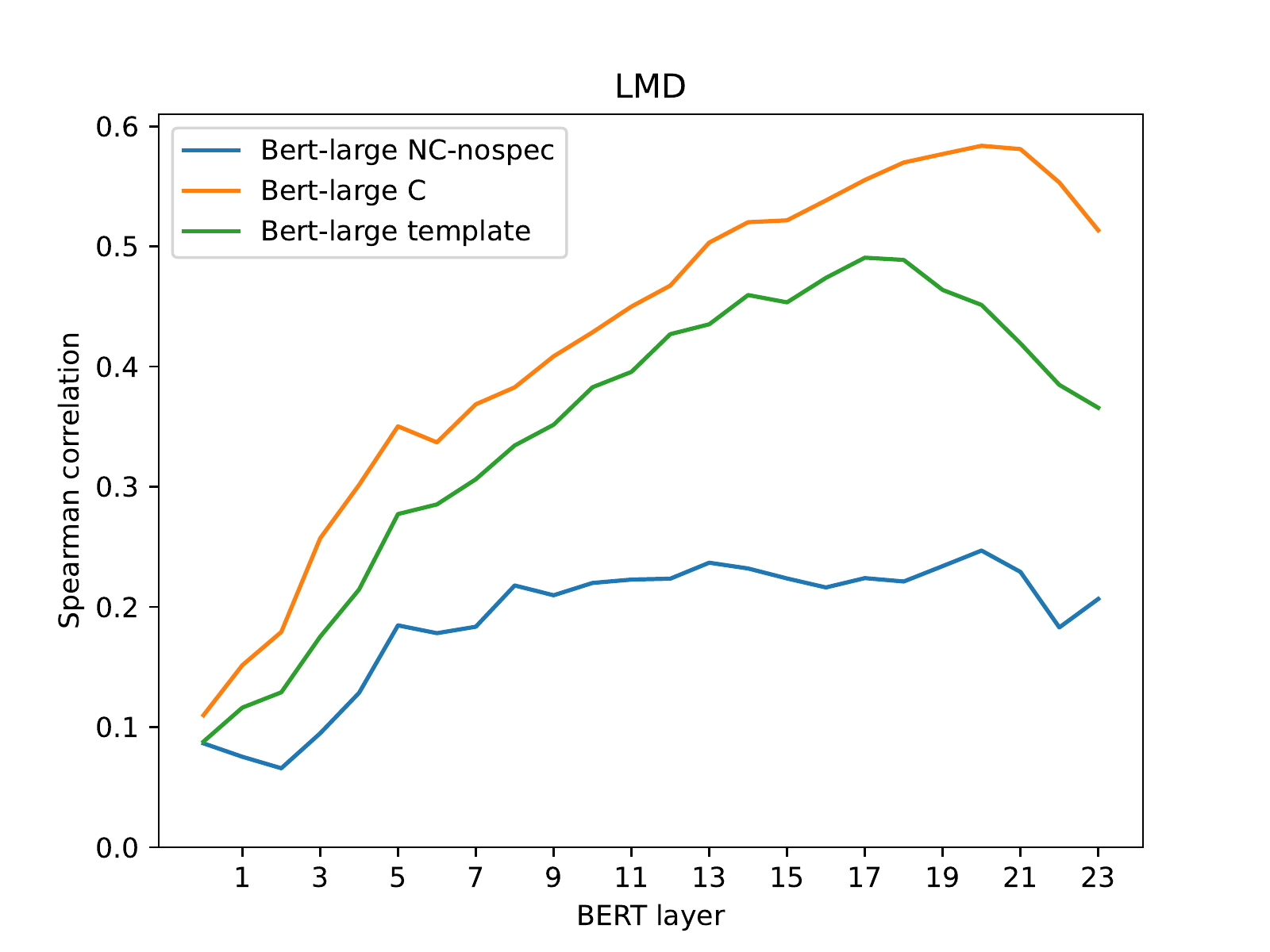}
    \caption{$\rho$ for LMD by BERT\textsubscript{large} \texttt{NC} (blue), \texttt{C} (orange), and \texttt{templated} (green). Best viewed in color.} \label{fig:LMD-temp}
\end{figure}

\begin{figure}[t!]
	\centering
	\includegraphics[width=1\linewidth]{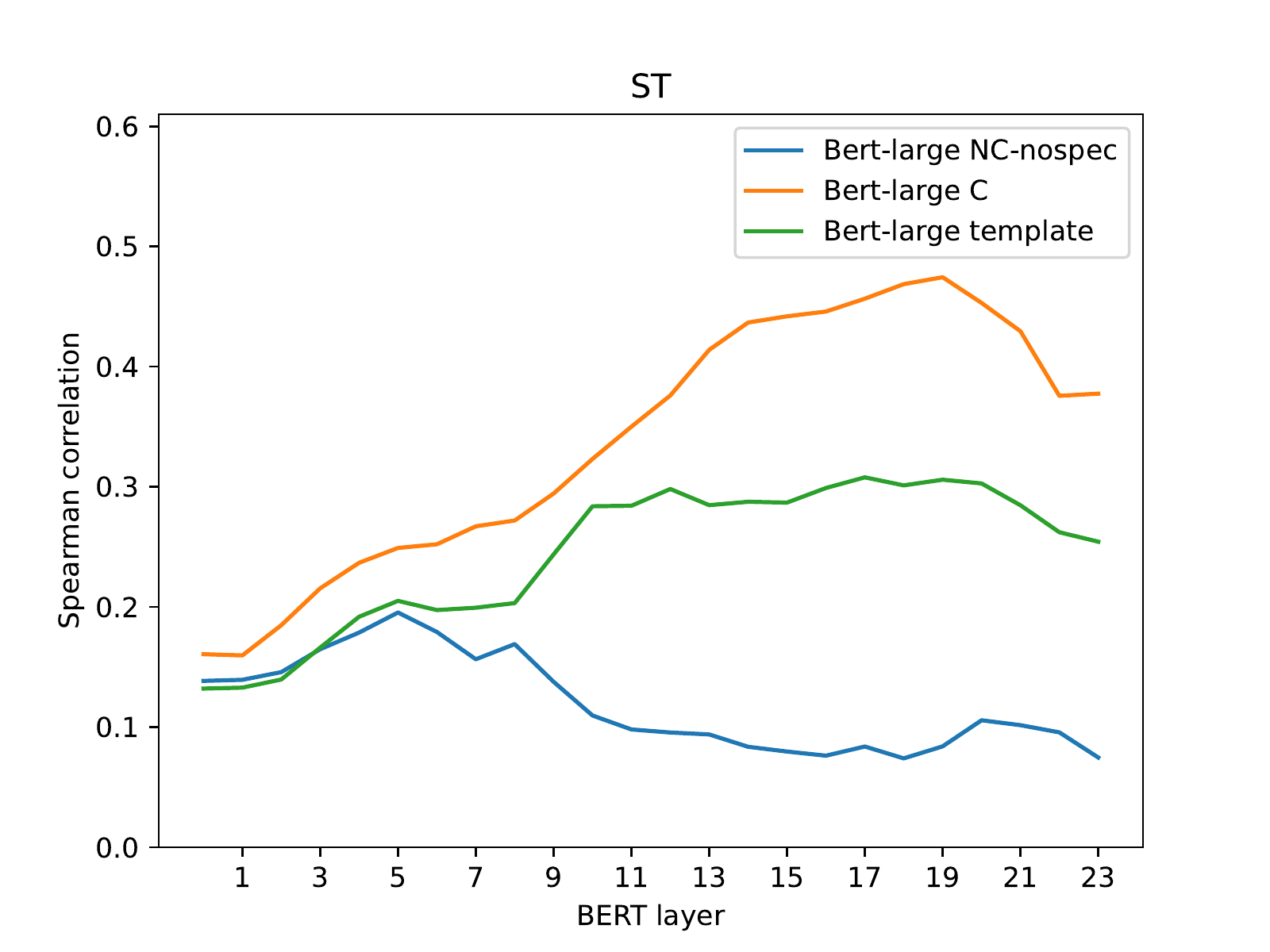}
    \caption{$\rho$ for ST by BERT\textsubscript{large} \texttt{NC} (blue), \texttt{C} (orange), and \texttt{templated} (green). Best viewed in color.} \label{fig:ST-temp}
\end{figure}

The results follow the expected pattern. As can be seen in Figure~\ref{fig:LMD-temp} and~\ref{fig:ST-temp}, the correlation values obtained in the \texttt{templated} setting (best $\rho$ for LMD: 0.491; best $\rho$ for ST: 0.308) lie somehow in between \texttt{C} and \texttt{NC}. While this shows that embedding words in a sentence leads to a representational advantage over the out-of-context presentation (they clearly outperform \texttt{NC}), these representations are still far behind \texttt{C} and either on par with (ST) or neatly below (LMD) the GloVe baseline.
This confirms that leveraging the meaningful linguistic context where compounds and constituents occur is crucial for obtaining sensible word representations that encode information on LMD and ST in line with human intuitions.

\subsection{Linear Regression Model}\label{sec:statistics}

\begin{figure}[t!]
	\centering
	\includegraphics[width=1\linewidth]{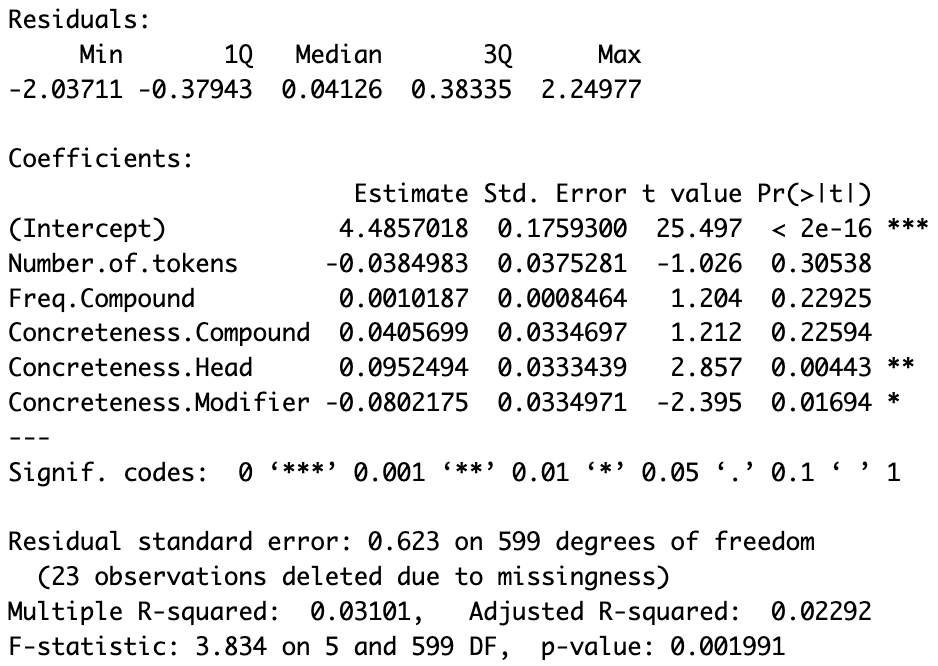}
    \caption{LMD. Linear regression model predicting LMD values by the best-performing \texttt{C} BERT\textsubscript{large} layer.} \label{fig:LMD-short}
\end{figure}

\begin{figure}[t!]
	\centering
	\includegraphics[width=1\linewidth]{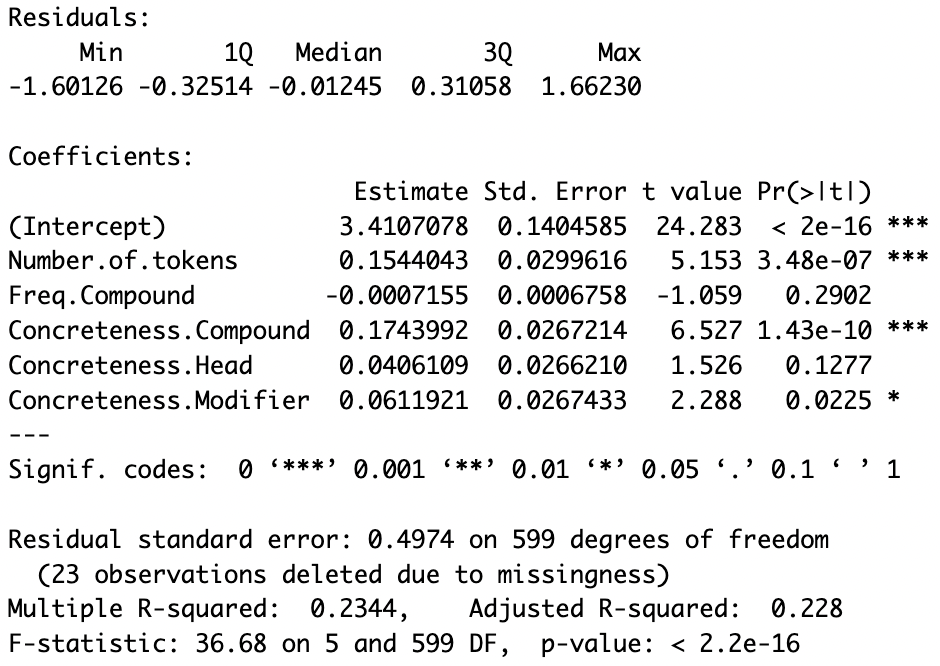}
    \caption{ST. Linear regression model predicting ST values by the best-performing \texttt{C} BERT\textsubscript{large} layer.} \label{fig:ST-short}
\end{figure}

Figure~\ref{fig:LMD-short} and~\ref{fig:ST-short} report all the effects and corresponding coefficients and p-values of the linear regression models described in Section~\ref{sec:R} for LMD and ST, respectively.

\end{document}